# When Classical Chinese Meets Machine Learning: Explaining the Relative Performances of Word and Sentence Segmentation Tasks

Chao-Lin Liu[1], Chang-Ting Chu, Wei-Ting Chang, Ti-Yong Zheng

Understanding classical Chinese is crucial for Chinese studies, partially because, in terms of time span, the majority of historical records about China were written in classical Chinese. Creating digital texts of the original documents either by optical record recognition or by human typists is required for employing digital assistance in Digital Humanities. The original Chinese texts usually did not include delimiters between words and sentences. Figure 1 shows a tomb biography that Liu and Chang used for discussing sentence segmentation of classical Chinese.[2] Unlike English, Chinese words are not separated by spaces, and there are no punctuation marks in the source texts. Word segmentation and sentence segmentation are thus important bedrocks for algorithmic understanding of classical Chinese, and the literature in digital humanities has started to see relevant discussions. Researchers applied statistical,[3,4] machine-learning,[5] or deep-learning[6] methods to train computational models for the segmentation tasks.

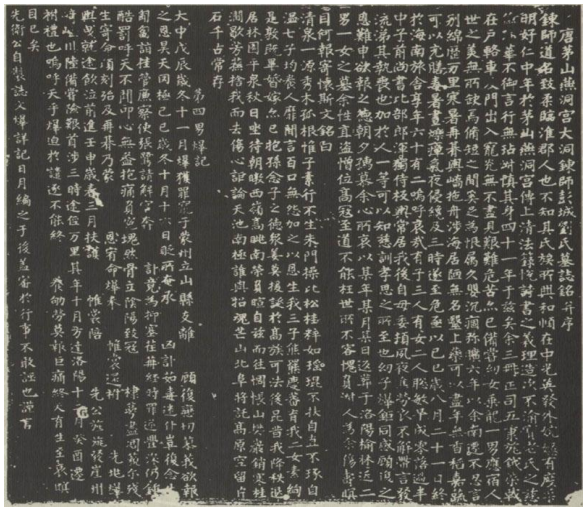

**Figure 1. A sample source of classical Chinese**

In standard machine-learning approaches, researchers acquire a collection of corpora for training and test purposes. Over the past many years, many, though not all, documents and books that were written in classical Chinese have been manually typed and saved as computer files. Modern punctuation marks have been added to the files at the same time. It is thus relatively easier to use the annotated texts for sentence segmentation than for word segmentation because it is very rare, if any, for typists to mark word boundaries in digitized texts. The training sets in Wang et al.'s, Han et al.'s and Shi et al.'s works have 237, 44, and 50 million characters, respectively.[7,8,9]

---

[2] Chao-Lin Liu and Yi Chang (2018) Classical Chinese sentence segmentation for tomb biographies of Tang dynasty, in *Proceedings of the 2018 International Conference on Digital Humanities*, 231–235.
[3] Ke Deng, Peter K. Bol, Kate J. Li, and Jun S. Liu (2016) On the unsupervised analysis of domain-specific Chinese texts, in *Proceedings of the National Academy of Sciences*, 113(22):6154–6159.
[4] Chao-Lin Liu (2019) Onto word segmentation of the *Complete Tang Poems*, in *Proceedings of the 2019 International Conference on Digital Humanities*, 2019.
[5] Ethem Alpaydin (2014) *Introduction to Machine Learning*, third edition, Cambridge: The MIT Press.
[6] Ian Goodfellow and Yoshua Bengio and Aaron Courville (2016) *Deep Learning*, Cambridge: The MIT Press.

Although we expect that using more training data should lead to better test results in general, this beneficial effect is not automatic. Due to the nature of machine-learning experiments, especially of deep-learning ones, it is not easy (or appropriate) to compare experimental results directly by their numbers. Various factors may influence the quality of segmentation. The best $F_1$ measure observed in Wang et al.'s work is slightly better than 80%.[7] When using a much larger training dataset for Qing, Han et al. did not report better results than when they used a much smaller dataset for the Tang epitaph.[8] Shi et al. defined their own measure for quality, and observed that the quality of segmentation varied between mid 60% and mid 90% in experiments for data of different Chinese dynasties.[9]

Machine-learning based methods provide us effective ways to build a running system for sentence segmentation. When provided with sufficient data, we may train computational models for sentence segmentation, and achieve impressive results.[2,7,8,9] Despite this convenience, previous research seldom aims at investigating the factors behind the success and perhaps failure of applications of deep-learning techniques for segmenting classical Chinese . Some pioneers had discussed the needs of explainable artificial intelligence as early as 2004,[10,11] and the demands are strengthening in recent years.[12] The quest for explanation not only satisfies our curiosity but also helps us build better systems for the segmentation tasks.

We may design different procedures to train computation models for sentence segmentation and observe the differences in the resulting quality of segmentation. The differences in the training procedures may provide clues as to what the computation models might learn from the training data. Though experimental findings alone are not solid proofs, the observations can be enlightening.

---

[7] Boli Wang, Xiaodong Shi, Zhixing Tan, Yidong Chen, and Weili Wang (2016) A sentence segmentation method for ancient Chinese texts based on NNLM. in *Proceedings of the Chinese Lexical Semantics Workshop* 2016, *Lecture Notes in Computer Science* 10085, 387–396.

[8] Xu Han, Hongsu Wang, Sanqian Zhang, Qunchao Fu, and Jun Liu (2019) Sentence segmentation for classical Chinese based on LSTM with radical embedding, *The Journal of China Universities of Posts and Telecommunications*, 26(2):1–8. (in Chinese)

[9] Xianji Shi, Kaiqi Fang, Xianjiong Shi, Xianju Shi, Xiandiao Shi, Xianji Shi, Xianda Shi, Xianfeng Shi, and Yanchun Song (2019) A method and implementation of automatic punctuation, *Journal of Digital Archives and Digital Humanities*, 3:1–19. (in Chinese)

[10] Michael van Lent, William Fisher, and Michael Mancuso (2004) An explainable artificial intelligence system for small-unit tactical behavior, in *Proceedings of the Sixteenth Conference on Innovative Applications of Artificial Intelligence*, 900–907.

[11] Mark G. Core, H. Chad Lane, Michael van Lent, Dave Gomboc, Steve Solomon, and Milton Rosenberg (2006) Building explainable artificial intelligence systems, in *Proceedings of Eighteenth Conference on Innovative Applications of Artificial Intelligence*, vol. 2, 1766–1773.

[12] Daniel S. Weld and Gagan Bansal (2019) The challenge of crafting intelligible intelligence, *Communications of the ACM*, 62(6):70–79.

ADHO DH2020

| Corpus | Number of Characters (NOC) | Number of Punctuations (NOP) | Ratio (NOP/NOC) |
|---|---|---|---|
| MZM (Tang Tomb Biographies, 唐代墓誌彙編) | 2421077 | 446937 | 18.46% |
| OTB (Old Tang Book, 舊唐書) | 1923361 | 359955 | 18.71% |
| NTB (New Tang Book, 新唐書) | 1559655 | 321131 | 20.59% |

**Table 1. Basic statistics of three corpora of classical Chinese (Tang dynasty)**

Table 1 provides some basic statistics about the corpora that were used in our experiments.[13] These corpora belonged to two genres and are representative documents of the Tang dynasty.[14] MZM contains only tomb biographies, and OTB and NTB are historical documents. To make the experimental results comparable, we split each of these corpora, and used 70% and 30% of each for training and test purposes, respectively.

We adopt a typical deep learning procedure for marking sentence boundaries in these corpora.[15] The corpora took turns to serve as training and test data. We produced the character embedding vectors of the corpora using Gensim,[16] and trained deep neural networks that are comprised of five layers of BiLSTM units in Keras.[17] We measured the quality of sentence segmentation with the $F_1$ measure.

Table 2 shows the observed $F_1$ measures for part of a system of experiments.[18] In the first experiment (ID=1), we used only MZM for training and test, and achieved 0.8772 in $F_1$. In Experiment 31, we used MZM, OTB, and NTB for training the LSTMs to segment the test portion of MZM. With the help of the extra training data, the $F_1$ increased slightly.

In Experiment 47, we produced character embedding with MZM, OTB, and NTB, trained the LSTMs with OTB to segment OTB, and achieved a better $F_1$ than that for Experiment 2. In Experiment 41, we added NTB to train the LSTMs, and improved the $F_1$. In contrast, in Experiment 35, we observed a smaller improvement when we added MZM to train the LSTMs. One possible reason for the reduction in improvement is that OTB is more similar with NTB than with MZM.

---

[13] Due to the word limit, we could not provide complete details about how we extracted the texts from the original documents to obtain these corpora. **MZM** is the complete contents of the two books of *Tang Tomb Biographies* (tang2 dai4 mu4 zhi4 hui4 bian1 唐代墓誌彙編) and one book of sequel (tang2 dai4 mu4 zhi4 hui4 bian1 xu4 ji2 唐代墓誌彙編續集). **OTB** contains the biographies (lie4 chuan4 列傳) in the *Old Tang Book* (jiu4 tang2 shu1 舊唐書), and **NTB** contains the biographies (lie4 chuan4 列傳) in the *New Tang Book* (xin1 tang2 shu2 新唐書).

[14] Tang is a Chinese dynasty that existed between 618CE and 907CE.

[15] Note that the step for word embedding and the step for training the LSTM units are separate.

[16] Gensim is available at https://radimrehurek.com/gensim/. When using *word2vec* to create the embedding vectors for Chinese characters, we set *window* (the window size) to 12, *min_count* to 1, *batch_words* to 8000, *iter* to 50, and produced vectors of 300 dimensions. We only report results of using the *skip-gram* algorithm.

[17] Keras is available at https://keras.io/. Inherited from our work that was reported in DH2019, we consider a context of six characters for determining whether a character is the last character of a sentence in training our BiLSTM stack. The outputs of the BiLSTM units have 400 dimensions, the optimizer is *Adam*, and we used *softmax* as the activation function for the final output layer. The output layer is a *Dense* layer, which used the *categorical_crossentropy* as the loss function.

[18] In Table 2, **M**, **O**, and **N** denote MZM, OTB, and NTB in Table 1, respectively. The columns M, O, and N indicate whether the corpora were used in creating word-embedding vectors and whether the corpora were used in training the LSTM stack. To facilitate the comparisons, we used only one corpus in the tests. We showed results for part of our experiments, so there are missing IDs in Table 2.

| ID | Word2Vec | | | Training LSTM Stack | | | Test | $F_1$ | ID | Word2Vec | | | Training LSTM Stack | | | Test | $F_1$ |
|---|---|---|---|---|---|---|---|---|---|---|---|---|---|---|---|---|---|
| | M | O | N | M | O | N | | | | M | O | N | M | O | N | | |
| 1 | M | | | M | | | MZM | **0.8776** | 40 | M | O | N | | O | N | MZM | 0.5858 |
| 2 | | O | | | O | | OTB | 0.7883 | 41 | M | O | N | | O | N | OTB | **0.8545** |
| 3 | | | N | | | N | NTB | 0.7907 | 42 | M | O | N | | O | N | NTB | 0.8246 |
| 31 | M | O | N | M | O | N | MZM | **0.8824** | 43 | M | O | N | M | | | MZM | 0.8769 |
| 32 | M | O | N | M | O | N | OTB | 0.8659 | 44 | M | O | N | M | | | OTB | **0.5695** |
| 33 | M | O | N | M | O | N | NTB | 0.8325 | 45 | M | O | N | M | | | NTB | 0.4964 |
| 34 | M | O | N | M | O | | MZM | 0.8822 | 46 | M | O | N | | O | | MZM | 0.5909 |
| 35 | M | O | N | M | O | | OTB | **0.8467** | 47 | M | O | N | | O | | OTB | **0.8325** |
| 36 | M | O | N | M | O | | NTB | 0.7480 | 48 | M | O | N | | O | | NTB | 0.7413 |
| 37 | M | O | N | M | | N | MZM | 0.8759 | 49 | M | O | N | | | N | MZM | 0.4881 |
| 38 | M | O | N | M | | N | OTB | 0.7778 | 50 | M | O | N | | | N | OTB | **0.7535** |
| 39 | M | O | N | M | | N | NTB | 0.8047 | 51 | M | O | N | | | N | NTB | 0.7921 |

**Table 2. Using different combinations of corpora to generate word vectors and train classifiers**

Another evidence for supporting this conjecture is that we achieved 0.7535 in Experiment 50 when we trained the LSTMs with NTB to segment OTB. When we used only MZM to train the LSTMs to segment OTB, the $F_1$ was only 0.5695 in Experiment 44. When we do not train the LSTMs with OTB to segment OTB, we achieved much better results when training the LSTMs with a more related corpus.

We have extended our work on sentence and word segmentation that were reported in DH2018 and DH2019. On the computing side, we have achieved better results by switching from traditional machine-learning methods to deep-learning techniques. As a study in Digital Humanities, we attempt to find qualitative explanations for the changing improvements which were influenced by the varying combinations of training data. We analyzed our observations based on results observed in a large number of experiments statistically. We could provide just one sample discussion for sentence segmentation tasks here, and we wish to have an opportunity to discuss more details about our experience in DH2020.

**The main body contains 994 words, excluding the title, the list of authors, the footnotes, the tables, and this statement.**